\documentclass[letterpaper, 10 pt, conference]{ieeeconf}  

\IEEEoverridecommandlockouts                              

\usepackage{graphics} 
\usepackage{epsfig} 
\usepackage{multicol}

\title{\LARGE \bf
Experience-Based Planning with Sparse Roadmap Spanners
}

\author{David Coleman$^{1}$, Ioan A. \c{S}ucan$^{2}$, Mark Moll$^{3}$, Kei Okada$^{4}$, and Nikolaus Correll$^{1}$
\thanks{DC was supported by the Google, JSK Lab at UTokyo, and NASA grant NNX12AQ47G. MM is supported in part by NSF NRI 1317849.}
\thanks{$^{1}$David Coleman and Nikolaus Correll are with the Department of Computer Science,
        University of Colorado, Boulder, CO 80309, USA
        {\tt\small [david.t.coleman|nikolaus.correll]@colorado.edu}}%
\thanks{$^{2}$Ioan A. \c{S}ucan is with Google [x],
        Mountain View, CA 94043, USA
        {\tt\small isucan@google.com}}%
\thanks{$^{3}$Mark Moll is with the Department of Computer Science, Rice University
        Houston, TX 77005, USA
        {\tt\small [mmoll@rice.edu]}}%
\thanks{$^{4}$Kei Okada is with JSK Lab, Univeristy of Tokyo
        Bunkyo, Tokyo 113-8656, Japan
        {\tt\small k-okada@jsk.t.u-tokyo.ac.jp}}%
}

\begin{document}

\maketitle
\thispagestyle{empty}
\pagestyle{empty}

\begin{abstract}
We present an experienced-based planning framework called \textit{Thunder} that learns to reduce computation time required to solve high-dimensional planning problems in varying environments. The approach is especially suited for large configuration spaces that include many invariant constraints, such as those found with whole body humanoid motion planning. Experiences are generated using probabilistic sampling and stored in a sparse roadmap spanner (SPARS), which provides asymptotically near-optimal coverage of the configuration space, making storing, retrieving, and repairing past experiences very efficient with respect to memory and time. The Thunder framework improves upon past experience-based planners by storing experiences in a graph rather than in individual paths, eliminating redundant information, providing more opportunities for path reuse, and providing a theoretical limit to the size of the experience graph. These properties also lead to improved handling of dynamically changing environments, reasoning about optimal paths, and reducing query resolution time. The approach is demonstrated on a 30 degrees of freedom humanoid robot and compared with the \emph{Lightning} framework, an experience-based planner that uses individual paths to store past experiences. In environments with variable obstacles and stability constraints, experiments show that Thunder is on average an order of magnitude faster than Lightning and planning from scratch. Thunder also uses 98.8\% less memory to store its experiences after 10,000 trials when compared to Lightning. Our framework is implemented and freely available in the Open Motion Planning Library.
\end{abstract}

\section{Introduction}

Most motion planning algorithms used in practice today are single-query planners~\cite{choset2005principles} that repeatedly solve similar problems from scratch without any re-use of past experience. A robot that is in operation for years will never get any better at its routine tasks. Even multi-query planners, which preprocess a roadmap of the environment, suffer from a high cost of construction and their inability to handle changing obstacles. The emerging field of experience-based planning aims at providing the best of both worlds, efficiently storing past solutions in memory to improve response speed and optimality of future requests.

\begin{figure}[!t]
\centering
\includegraphics[width=3.4in]{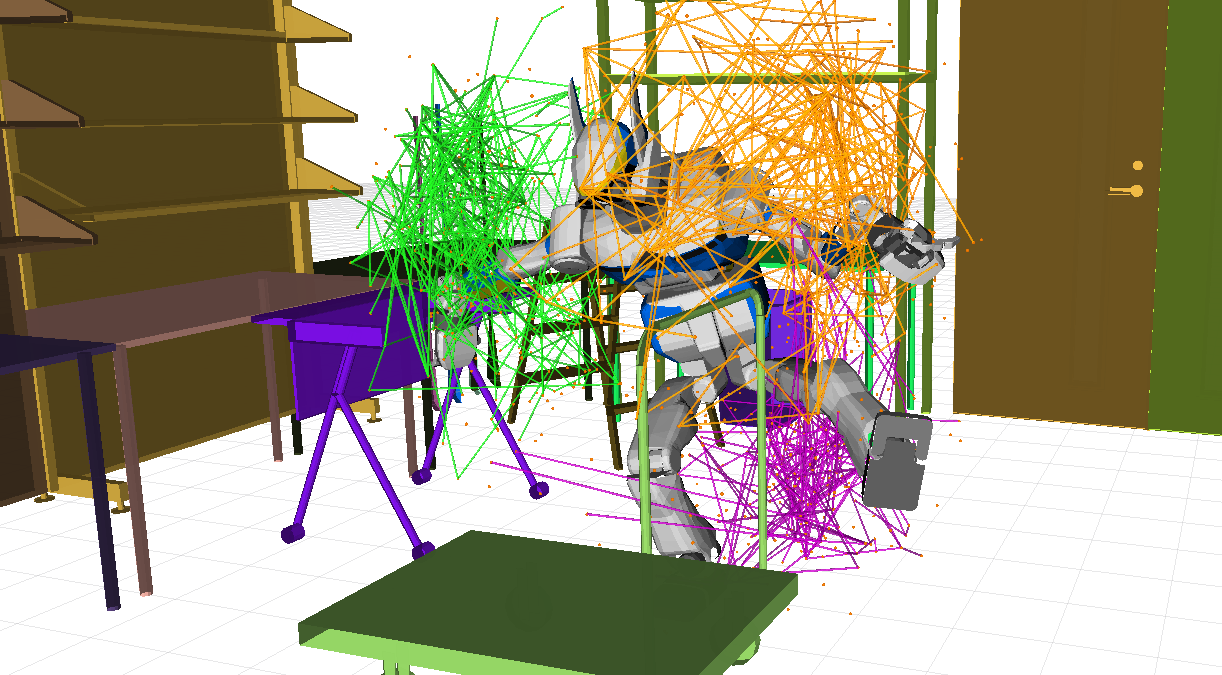}
\caption{Visualization of HRP2's experience database after 10,000 runs with obstacles. The three different colored line graphs project a whole body state to a reduced image of each of the robot's two end effectors and free foot.}
\label{fig:hrp2_db}
\end{figure}

Efficiently saving and recalling past solutions poses a series of challenges. As the dimensionality of the planning problem increases, the necessary size of a naive cache implementation becomes infeasible. Determining how many experiences should be saved and in which format to save them is one problem. Recalling and identifying the best experience --- if one exists --- is another issue that becomes especially hard when there are varying constraints and collision environments. Finally, efficiently repairing a partial solution from recall to work with the current planning problem is a third challenge.  As a robot moves from different tasks, such as working in a kitchen versus climbing a ladder, it is important that changing obstacle locations can efficiently be considered when recalling a path. These problems become particularly challenging when considering not only kinematics, but also dynamics, which at least doubles the dimensionality of the search space and imposes additional constraints. In this paper only kinematic planning is addressed.
 
Experienced-based planning is especially suited for robots with a large number of invariant constraints, such as joint limits, self-collision, and stability constraints because these expensive checks can be inherently encoded in the experience database. Whenever a path is recalled, it is guaranteed to already satisfy the invariant constraints. Variant constraints are mainly collision checking with obstacles. An example of a highly invariant-constrained robot is a biped humanoid, which although kinematically has a large range of motion, has only a limited free space that satisfies quasi-static stability.

To build an efficient experience database, it is our goal to have a roadmap in body coordinates that can capture all past homotopic classes a robot has experienced throughout various collision environments. This means that for difficult path planning problems such as the narrow passageway problem, at least one path is saved that traverses through this constrained region, and it can be later recalled and continuously deformed into a near-optimal solution. 
%
\subsection{Related work}

Experience-based planning has its roots in Probabilistic Road Maps (PRM) \cite{prm}, a multi-query path planning algorithm that stores pre-computed collision-free trajectories in a graph. A key limitation of PRMs is that the resulting roadmaps have no limits to their size and can contain information that the robot might never require. 

An alternative approach is to store only paths that the robot has actually used \cite{lightning_framework,Stolle_2007_6097}. These methods map a current query to an exact past path that can be quickly recalled, similar to a lookup table. Path-centric approaches suffer from limited dataset sizes or conversely, too much memory usage and they also lack generalization of experiences.   

Saving past experiences in graphs, the subject of this paper, has been explored in \cite{phillips2012graphs} with their E-Graphs approach. However, E-Graphs use discrete representations of the world and low-dimensional projection heuristics that allow the planner to get stuck in local minima. E-Graphs use task-based requests and they map poorly to robots with complex high-DOF kinematics. In addition to similar expanding memory challenges as other approaches, E-Graphs require that the entire graph be collision checked before each query.

Other approaches include recalling attractors to bias sampling in areas of higher probability \cite{attractors}, using motion primitives in probabilistic sampling-based planning \cite{hauser2008using}, and caching configuration space approximations for improving randomized sampling on constraint manifolds \cite{sucan2012motion}.

A major challenge in any experience-based planner is to retrieve and repair, or adapt, past experiences. This topic is still an active area of research \cite{agarwal2011compact} and as such no consensus as to the best method for performing this exists. In Lightning \cite{lightning_framework} the most relevant and similar experience are retrieved using two heuristics: the top $n$ similar candidate paths are found by searching for the path with the closest pair of both start and goal states. The second heuristic scores the $n$ nearby paths by calculating the percent of the path ($pscore$) that is now invalid due to variant constraints such as changing obstacles. The nearby path with the lowest $pscore$ is chosen as the path to be ``repaired''. The repair step uses a bidirectional RRT to attempt to re-connect each set of end points of segments that have been disconnected by invalid regions. In many cases the computation of recalling and repairing a path, in particular difficult paths containing challenges like a narrow passage~\cite{narrow_passage}, is less than the computation required for planning from scratch~\cite{lightning_framework}.

\begin{figure}[!t]
\centering
\includegraphics[width=3.4in]{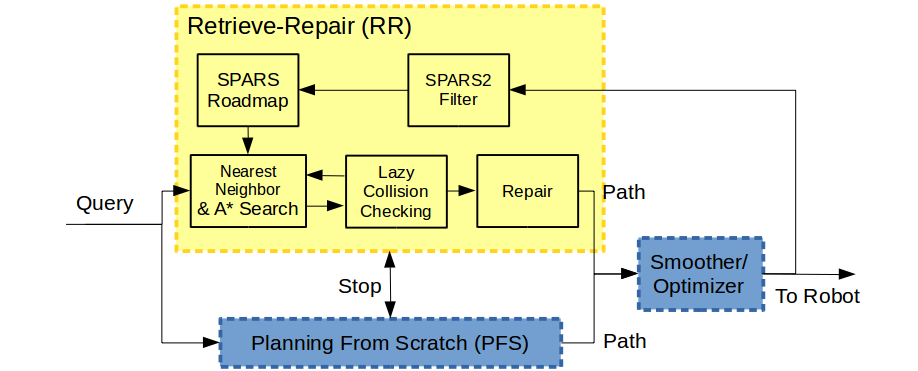}
\caption{Diagram of the Thunder framework}
\label{fig:thunder_diagram}
\end{figure}

\subsection{Contribution of this paper}
In this paper, past experiences are stored in a sparse roadmap spanner \cite{spars1,spars2}, which allows us to combine the advantages of connected roadmaps, namely generalizing from past experiences, with the attractive memory and lookup properties of sparse roadmap spanners. Our framework unifies several approaches and it is tested on a whole body humanoid in variable obstacle environments.

\section{The Thunder Framework}

The Thunder Framework is built on the parallel module concept of the Lightning Framework \cite{lightning_framework}. As in \cite{lightning_framework}, computation is split into two modules. The first module runs ``planning from scratch'' (PFS), which attempts to solve the problem using a traditional sampling-based planner without prior knowledge or experience. The second thread runs ``retrieve and repair'' (RR), which uses an experience database to find a past solution that is similar to the current problem, and then attempts to repair that solution as necessary to solve the current problem. The solution from the component first to finish is returned as the overall result, while the other module is immediately terminated, as shown in Figure~\ref{fig:thunder_diagram}. 

The PFS component maintains the same guarantee of probabilistic completeness as a traditional sampling-based planner \cite{choset2005principles}. It is used to initialize the empty experience database so that no human training is required, allowing a robot, which typically has different kinematics and joint limits than a human, to find solutions that would be difficult to specify for human trainers.

Our approach to store past experiences builds upon sparse roadmap spanners (SPARS) \cite{spars1} that compactly represent a graph while maintaining the original optimality to within a stretch factor $t$. The SPARS algorithm is powerful in that it is probabilistically complete, asymptotically near-optimal, and the probability of adding new nodes to the roadmap converges to zero as new experiences are added. Previous works similar to SPARS suffered from unbounded growth~\cite{prm_star}, or the inability to only remove edges but not nodes~\cite{marble2011asymptotically}. SPARS uses graph spanners to create subgraphs where the shortest path between two nodes is no longer than the stretch factor $t$ times the shortest path on the original graph. This allows theoretical guarantees on path quality to be upheld while filtering out unnecessary nodes and edges from being added to the graph. 

In order to have the same asymptotic optimality guarantees as PRM*, within a $t$-stretch factor, a number of checks are required to determine which potential nodes and edges (experiences) should be saved to have coverage across a robot's free space. Only configurations that are useful for 1) coverage, 2) connectivity, or 3) improving the quality of paths on sparse roadmap relative to the optimal paths in the configuration space. Two parameters $t$ and the sparse delta range $\Delta$ control the sparsity of the graph. 

This paper uses an improved version, called SPARS2 \cite{spars2}, which relaxes the requirement for a dense graph to be maintained alongside the sparse graph. This greatly reduces the memory requirements of the graph, making it practical for high DOF configuration spaces to be easily maintained in memory. 

Collision checking is one of the most computationally intense components of most motion planners, and as such reducing the amount of required collision checks is pivotal in increasing speed. Lazy collision checking, as demonstrated in LazyPRM \cite{lazyprm} is a classic approach used in Thunder that delays checking until after a candidate path through a configuration space is found. Once a path is found, it is checked for validity against the changing collision environment. Invalid segments are removed or marked as invalid, and search continues.

\subsection{Recording experiences} \label{sec:inserting}

\begin{figure}[!t]
\centering
\includegraphics[width=3.4in]{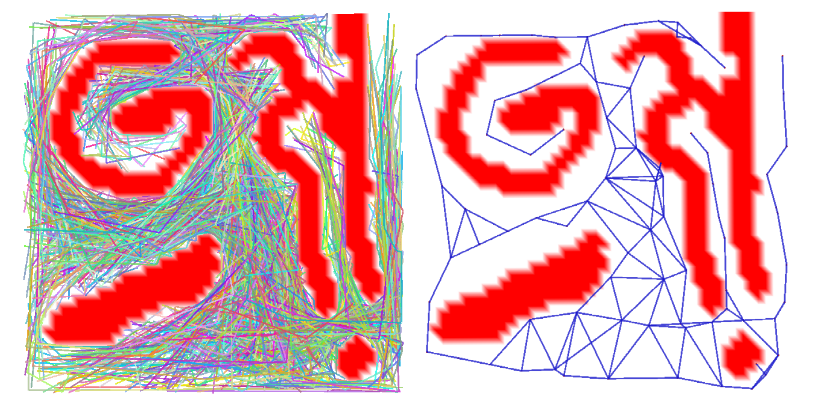}
\caption{Visualization of 2000 experiences in a 2D world saved in A) Lightning and B) SPARS}
\label{fig:graph_approaches}
\end{figure}

In our approach, a robot's experience database is initially empty. As solutions are generated from the PFS planner, they are fed back into the experience database. Every solution path is discretized into an ordered set of vertices and incorporated in the experience graph. SPARS' theoretical guarantees decide which parts of past experiences to save. 

In practice, care must be taken when incorporating a discretized path into an experience roadmap while maintaining the properties of SPARS. The difficulty stems from the SPARS checks that each inserted node must pass. The naive approach of inserting a new path into the experience database linearly --- inserting each node in order from start to goal --- will almost never yield a single connected component in the graph, early on.

\begin{figure}[!t]
\centering
\includegraphics[width=3.4in]{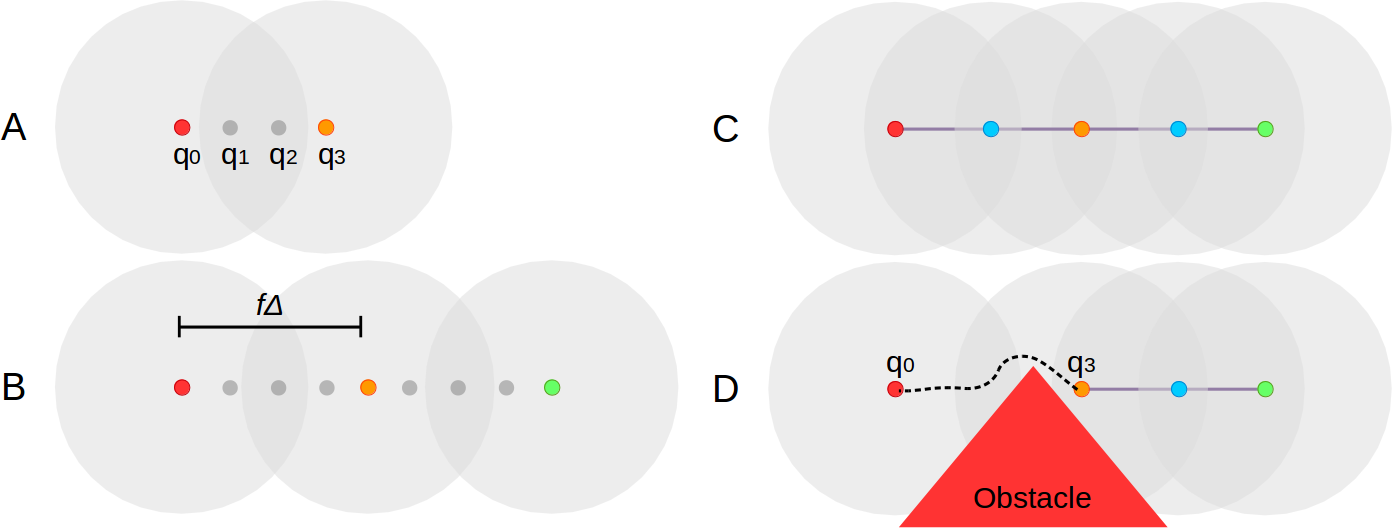}
\caption{Node types: start (green), goal (red), \textit{coverage} guards (orange), \textit{connectivity} guards (blue). Grey circles indicate visibility regions and the dotted line indicates a disconnected segment. A) demonstrating in-order insertion that fails to create an edge. B) evenly spaced out guard placement. C) guards are connected with edges D: potential issue where spacing does not work out with constraints in configuration space.}
\label{fig:simple_path}
\end{figure}

As a simple example, take an empty graph in a configuration space devoid of obstacles and a discretized candidate path, as shown in Figure \ref{fig:simple_path}A. Initially a node $q_0$ is attempted to be inserted in a free space. Because there are no other nearby nodes, it is added for coverage. The next nodes, $q_1$ and $q_2$, in the ordered set of discretized path states will likely still be within the sparse delta range $\Delta$ of $q_0$ and will be visible to $q_0$ because there are no obstacles. These nodes are not added because SPARS determines that $q_0$ can serve as their representative guard. Next, node $q_3$ will be attempted to be inserted into the graph spanner. Because it is no longer visible to $q_0$, it will also be added as a guard of type \textit{coverage}. However, no edge will be created connecting coverage nodes $q_0$ to $q_3$ because edges are only added when 1) a candidate node is visible to two other surrounding nodes that are not already part of the same connected component, or 2) when they are on an \textit{interface} or border between the visibility regions of two guards. From this simple demonstration, a graph of two disconnected components is created that fails to bridge a connected path.

Having a disconnected path is not detrimental because future similar paths have a probability of eventually reconnecting the components with their candidate states. In the meantime, however, the retrieval component of Thunder suffers from a large number of disconnected components. The size of a high-dimensional configuration space can make the probability of two similar paths being inserted rare. Therefore, it is advantageous to ensure that a candidate path is fully inserted and connected in order to greatly increase the rate at which a Thunder experience database becomes useful. 

\begin{figure*}[!t]
\centering
\includegraphics[width=\textwidth]{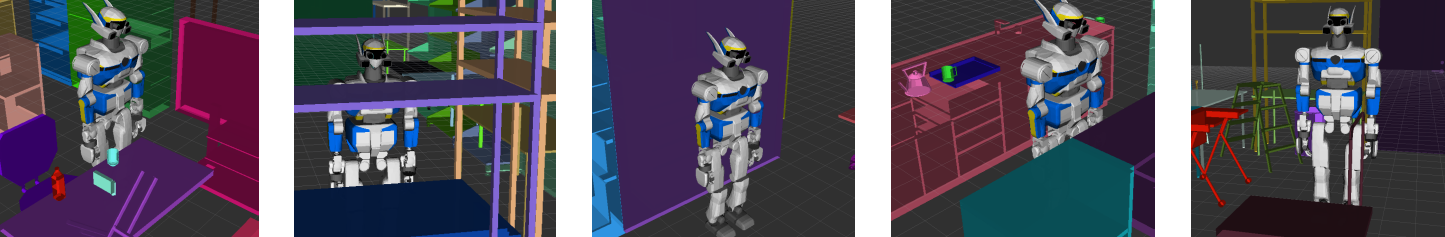}
\caption{The five collision environments for HRP2 used to test repairing experiences}
\label{fig:collision_environments}
\end{figure*}

To improve connectivity of inserted paths, a straight-path heuristic that orders the insertion of interpolated states into SPARS is utilized. The heuristic is built on the premise that two \textit{coverage} guards must be within a certain distance $d$ of each other. That distance is $\Delta \cdot f_{low} < d < \Delta \cdot f_{high}$ where the sparse delta scaling factor ranges $f$ must be between 1.0 and 2.0 as required by SPARS' visibility criteria, shown in Figure \ref{fig:simple_path}B. The best value for $f$ for our candidate path is chosen such that guards are evenly spaced out to prevent any of the \textit{coverage} guards from being inserted:

\begin{multicols}{2}
   \noindent
  $$f = \frac{l}{n \cdot \Delta}$$
  $$n = \lfloor \frac{l}{\Delta \cdot f_{low}} \rfloor$$
\end{multicols}

\noindent
where $l$ is the total length of the path, $n$ is the number of guards, and the second relation is rounded down. Using this result, a guard is inserted every $f\cdot\Delta$ distance away from the last guard and have an even coverage of the path, as shown in Figure \ref{fig:simple_path}B. To then connect those \textit{coverage} guards with edges, new nodes must be inserted in between each pair of \textit{coverage} guards to create the necessary \textit{connectivity} guards. For a straight line path with no obstacles, a graph like the one shown in Figure \ref{fig:simple_path}C will emerge.

In the aforementioned insertion method, it was assumed that the path was straight and there were no obstacles in the environment. However, adversarial cases such as when a path has a small curve (dotted line) around an obstacle, as shown in Figure \ref{fig:simple_path}D, can cause a path to become disconnected. The utilized $f$ in this example is not able to capture this curve because of the chosen resolution, and no edge will connect $q_0$ to $q_3$ because they lack visibility. Similar scenarios exist even without obstacles when paths are convoluted, such as a zig-zag shape.

To mitigate for these edge cases, all remaining path states that were not already attempted to be added to the sparse roadmap are added in random order. While still no guarantees exist for full connectivity, the probability is much higher since an informed insertion heuristic has already achieved most of the necessary connections.

The path discretization resolution was set as the same used in collision checking, which results in many unnecessary node insertion attempts but greatly improves the chances of a fully connected path being added to the database. In practice, a path that is attempted to be inserted into a sparse roadmap achieves start-goal connectivity about 97\% of the time.

\subsection{Retrieving and Repairing Experiences} \label{sec:retrieving}

Finding a valid experience in the Thunder experience database is similar to the standard PRM and SPARS approach of using A* to search for the optimal path in the experience roadmap. The main difference is that lazy collision checking is utilized so that the framework can handle changing collision environments.

Our search method is as follows. For both the start and the goal states $q_{start}$ and $q_{goal}$, all saved states within a $\Delta$ radius are found. Every combination of pairs of nearby candidate start/goal nodes $q_{nearStart}$ and $q_{nearGoal}$ are checked for visibility to their corresponding $q_{start}$ and $q_{goal}$ states. If the start and goal are successfully connected to the graph, A* searches over the graph to find an optimal path. In some cases the roadmap contains disconnected components, in which case a path may not be found. However, in our experiments even with a 30 DOF robot this is rarely the case.

Once a path is found, each edge is checked for collision with the current environment. Although the path is guaranteed to be free of invariant constraints it is necessary to still check that no new obstacles exist that may invalidate our path. If a candidate solution is found to have invalid edges, those edges are disabled and A* searches again until a completely valid path is found, or it is determined that no path exists.

If a valid path is found, the path is smoothed and sent to the robot for execution. If no path is found, other pairs of candidate start/goal nodes $q_{nearStart}$ and $q_{nearGoal}$ are attempted. If still no path is found, the candidate path with the fewest invalid segments is repaired, if there exists one. To repair the invalid path, the repair technique used in the Lightning Framework is employed, such that a bi-directional RRT attempts to reconnect disconnected states along the candidate path. If the PFS module finds a solution before a path can be retrieved and repaired, the recall request is canceled.

\section{Results}

We ran a series of benchmarks to determine the efficacy of the Thunder Framework using the Open Motion Planning Library \cite{ompl} and MoveIt! \cite{coleman2014reducing} with a rigid body humanoid. All tests were implemented in C++ and run on a Intel i7 Linux machine with 6 3.50GHz cores and 16 GB of RAM.

First, we implemented and benchmarked the original Lightning Framework for a baseline comparison. The parameters of Lightning were tested and tweaked, optimizing for performance: the dynamic time warping threshold used for scoring path similarity was set to 4 and the number of closest candidate paths to test 10 in all experiments. Thunder was then implemented using a modified version of SPARS2 as our experience database with the SPARS2 $\Delta$ fraction set to 0.1 and $t$ set to 1.2. The lazy collision checking module was integrated and the insertion and recall methods incorporated as described in Section \ref{sec:inserting} and \ref{sec:retrieving}. 

PFS, Lightning, and Thunder frameworks were compared. For all three methods the bi-directional planner RRTConnect~\cite{rrtconnect} was used to facilitate comparison with the original Lightning implementation. Scratch planning was tested using two threads for a fair comparison with the experience frameworks, which both use two threads.
 
MoveIt! was modified to allow whole body motion planning for bipeds such as the 30 DOF HRP2 humanoid. Although some preliminary experiments were conducted on hardware, the results presented here are from simulation. The robot was placed in 5 different collision environments to test its ability to repair past invalid experiences, as shown in Figure~\ref{fig:collision_environments}.

A single contact point, in our case the robot's left foot, is fixed to the ground and the rest of the robot balances on that foot. During PFS, all of the humanoid's joints are sampled randomly and each candidate robot state is checked for stability, self-collision, and collision with the environment. Quasi-static stability is checked by maintaining the center of mass within the foot's support polygon. The support polygon was reduced by 10\% to account for modeling and mechanical error.

\begin{figure}[!t]
\centering
\includegraphics[width=3.4in]{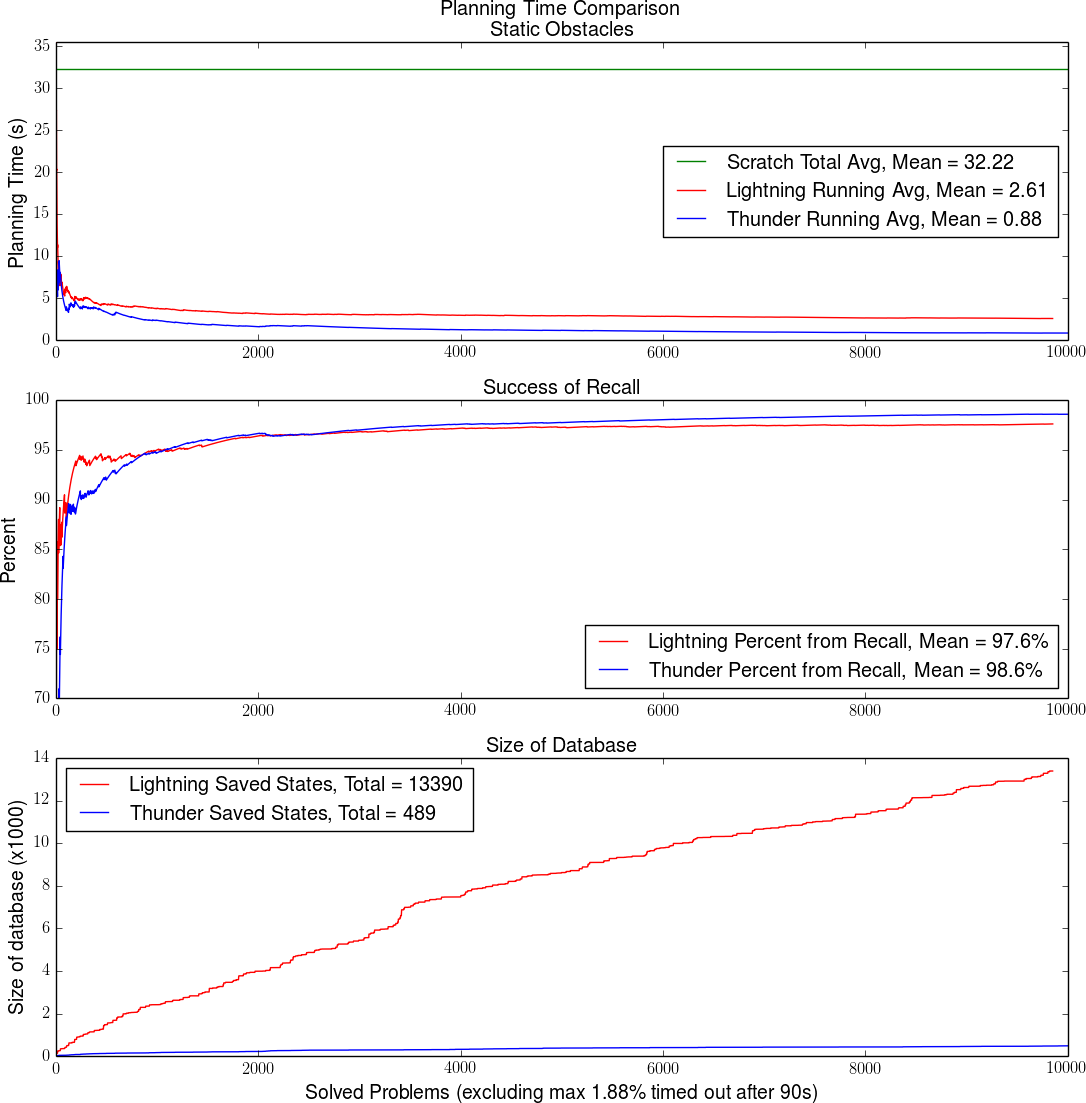}
\caption{Comparison of methods with a static collision environment.}
\label{fig:result_with_obstacles}
\end{figure}

In the first test the performance of experience planning was tested in a static, unchanging collision environment. The start state was always the same, and the goal state random. Planning timed out after 90 seconds and problems that had not been solved after that time are discarded, which accounted for less than 1.88\% of runs for each method. Ten thousand runs were performed to observe how the database grew over time. Results comparing planning time, frequency of using recalled paths, and growth of the databases are shown in Figure~\ref{fig:result_with_obstacles}.

The second test makes the planning problem more realistic and difficult by randomly choosing an obstacle environment for each planning problem from the five shown in Figure~\ref{fig:collision_environments}. This increased difficulty resulted in 3.02\% of runs for each method being discarded due to timeouts. Results after 10,000 motion plan trials are shown in Figure~\ref{fig:result_rand_obstacles}.

We also compare the size difference between the experience databases in Thunder and Lightning in the second experiment. Thunder's database uses 235 kB and Lightning uses 19,373 kB, meaning Thunder uses 98.8\% less storage than Lightning. Similarly, Thunder had stored 621 states, and Lightning 58,425 states. Finally, the variance in response time is shown in Figure~\ref{fig:result_rand_obstacles_hist} for the three methods.

The average insertion time of new paths into the databases was 1.41 seconds for Thunder and Lightning 0.013 seconds for Lightning. Path insertions are performed in the background and do not affect planning time.

\section{Discussion}

\begin{figure}[!t]
\centering
\includegraphics[width=3.4in]{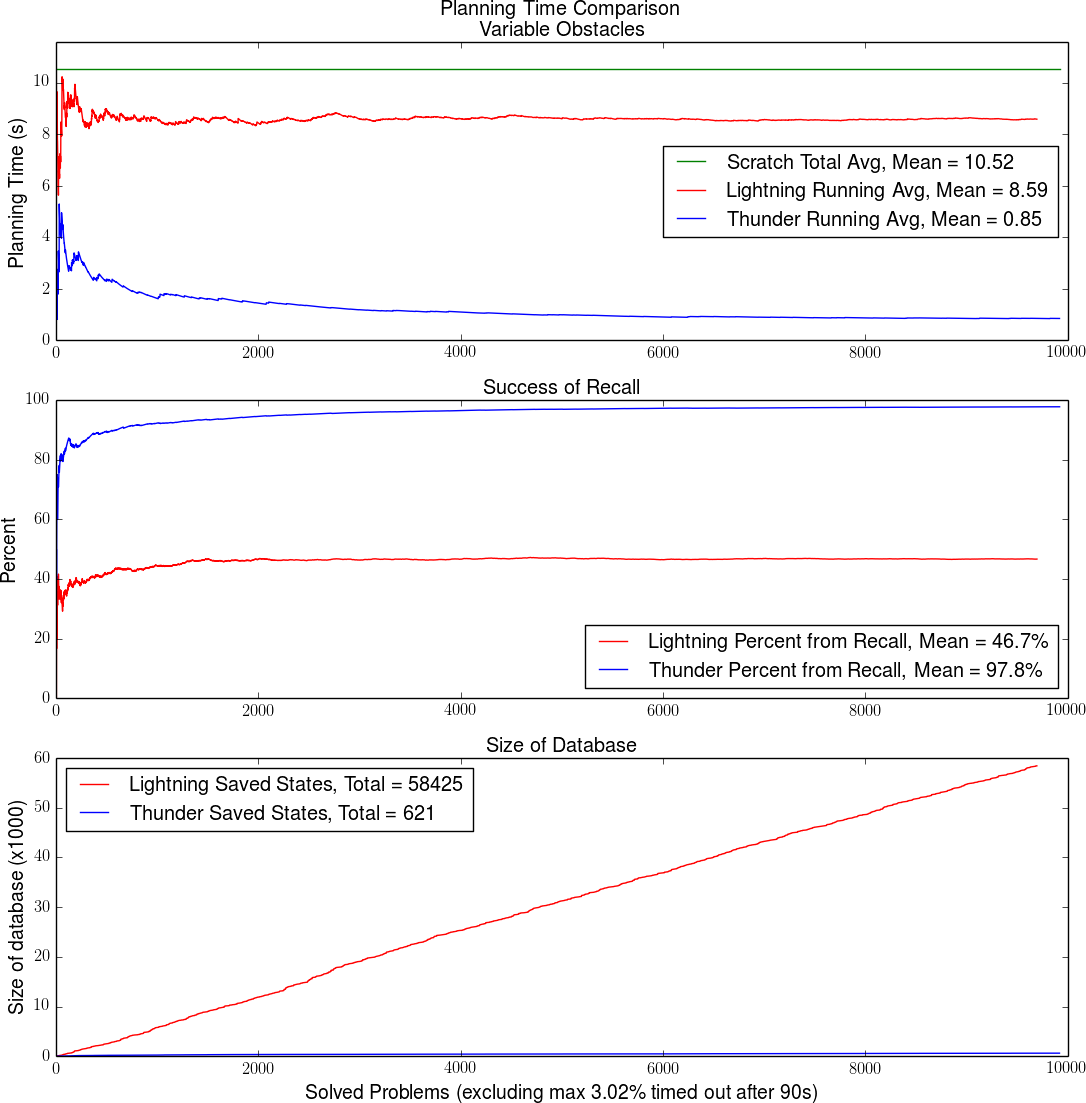}
\caption{Comparison of methods with varying collision environments.}
\label{fig:result_rand_obstacles}
\end{figure}

Our results demonstrate that planning from past experience, using either of the Thunder or Lightning approaches, provides vastly improved query response times in hard motion planning problems. The difference is most notable in problems with a large amount of invariant constraints, such as in the top of Figure \ref{fig:result_rand_obstacles}. From this we see that after 10,000 runs Thunder outperforms Lightning by a factor of 10.0 and PFS by a factor of 12.3. Thunder on average solves problems 1231\% faster than PFS. However, scratch planning can sometimes be faster for simple problems such as when the start and goal states are nearby. 

\begin{figure}[!t]
\centering
\includegraphics[width=2.95in]{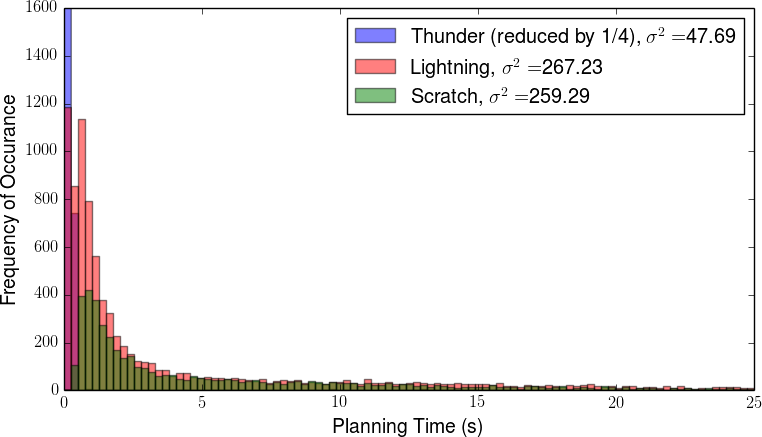}
\caption{Histogram of planning time across 10,000 tests. Note that Thunder frequency has been reduced by 1/4 to improve readability.}
\label{fig:result_rand_obstacles_hist}
\end{figure}

The variance in response time also improves with a graph-based experience database, as shown in Figure \ref{fig:result_rand_obstacles_hist}. A path-centric experience approach requires far more repair planning using traditional sampling methods, which decreases the deterministic response time of the path. Using Thunder after it was properly trained also has the added benefit of returning paths that are more predictable for a repeated query.

In addition to response time, memory usage is also vastly improved in our approach. Because a path-centric approach is unable to reuse subsegments, it is forced to store many similar and redundant paths which result in huge amounts of wasted memory. A path-centric experience planner suffers from the ability to only connect at the start and end of its stored paths. After 10,000 runs of random start-goal planning problems, the Lightning framework had grown to a database two orders of magnitude (98.8\%) larger than Thunder's and it appears that it will continue to increase linearly with time despite having a path similarity filter.

The SPARS theoretical claim that the probability of adding new nodes goes to zero with time have been empirically verified; during the last 1,000 trials of the second test, new nodes were being added to the Thunder database at a rate of 0.03 states per problem while Lightning was still increasing at a rate of 5.61 states per problem.

Another performance difference observed between frameworks was the frequency that a recalled path was used as shown in the middle graph of Figure \ref{fig:result_rand_obstacles}. In Thunder, after building a sufficient experience database, the number of problems solved by recall was 96.7\%. In contrast, Lightning remained at a recall rate of 46.9\% despite having a much larger database, indicating that the heuristics used in finding good paths to repair could be improved.

It was observed during experimentation that smoothing of PFS paths is important before inserting an experience into the database to reduce the chances that unnecessary curves break connectivity of an inserted path. 

One major drawback from using SPARS as our database is the dramatic difference in insertion time. A solution path must first be finely discretized into a set of states, and each state must be tested against the graph spanner properties discussed in \cite{spars1}. This verification step is computationally intensive, and therefore to save time, in Thunder only experiences that were planned from scratch are candidates for insertion. Still, the average insertion time of Lightning is two orders of magnitude  faster than Thunder. We do not include this computational time in our benchmarks however, because this step can easily be done offline, or while the robot is idle.

Another drawback to Thunder over PFS is that often once one solution, or homotopic class, is discovered and saved for a problem, shorter or lower cost paths can be overlooked in future, similar queries. This is because better solutions can only be discovered by PFS, and typically the recalled path will be found first and the scratch planning will be canceled. This problem is most common when the free configuration space changes, such as an obstacle being removed. 

\section{Conclusion}

Replanning without taking advantage of past knowledge has been demonstrated to be less computationally efficient than leveraging experience-based planning. Our approach of using SPARS2 has shown that much less memory can be used in saving experiences for large configuration spaces. This has the added benefit that multiple experience databases can easily be maintained in memory, allowing task or environment-specific graphs to be stored and recalled. There are many areas of future work including improving the aforementioned insertion time, the ability to discover shortcuts that appear with changing obstacles, and the ability to handle dynamics in non-holonomic systems. Additionally, smarter collision checking, choosing the best path discretization, finding the best SPARS stretch factor, and implementing massive parallelization are areas of further investigation. Finally, accounting for optimization objectives is an area of particular interest.

\addtolength{\textheight}{-12cm}   



\bibliographystyle{IEEEtran} 
\bibliography{thunder}

\end{document}